\documentclass{article}
\usepackage{ijcai22}

\usepackage{times}
\usepackage{soul}
\usepackage{url}
\usepackage[hidelinks]{hyperref}
\usepackage[utf8]{inputenc}
\usepackage[small]{caption}
\usepackage{graphicx}
\usepackage{amsmath}
\usepackage{amsthm}
\usepackage{booktabs}
\usepackage{algorithm}
\usepackage{algorithmic}
\usepackage{multirow}
\usepackage{subfigure}

\title{Robust Single Image Dehazing Based on Consistent and Contrast-Assisted Reconstruction}

\author{
De Cheng$^1$
\and
Yan Li$^2$\and
Dingwen Zhang $^{3}$\and
Nannan Wang $^1$\and
Xinbo Gao $^4$\and
Jiande Sun $^2$
\affiliations
$^1$Xidian University,
$^2$Shandong Normal University\\
$^3$Northwestern Polytechnical University,
$^4$Chongqing University of Posts and Telecommunications
}

\begin{document}

\maketitle


\begin{abstract}
  Single image dehazing as a fundamental low-level vision task, is essential for the development of robust intelligent surveillance system. In this paper, we make an early effort to consider dehazing robustness under variational haze density, which is a realistic while under-studied problem in the research filed of singe image dehazing. To properly address this problem, we propose a novel density-variational learning framework to improve the robustness of the image dehzing model assisted by a variety of negative hazy images, to better deal with various complex hazy scenarios. Specifically, the dehazing network is optimized under the consistency-regularized framework with the proposed Contrast-Assisted Reconstruction Loss (CARL). The CARL can fully exploit the negative information to facilitate the traditional positive-orient dehazing objective function, by squeezing the dehazed image to its clean target from different directions. Meanwhile, the consistency regularization keeps consistent outputs given multi-level hazy images, thus improving the model robustness. Extensive experimental results on two synthetic and  three real-world datasets demonstrate that our method significantly surpasses the state-of-the-art approaches.


\end{abstract}


\section{Introduction}
Image dehazing aims to recover the clean image from a hazy input, which is essential for the development of robust computer vision systems. It helps to mitigate the side-effect of image distortion induced by the environmental conditions, on many visual analysis tasks, such as object detection~\cite{li2018end} and scene understanding~\cite{sakaridis2018semantic}. Therefore, single image dehazing has attracted more and more attention, and many dehazing methods have been proposed recently~\cite{qin2020ffa,wu2021contrastive}.

\begin{figure}[t!]
	\vspace{-4mm}
	\scriptsize
	\centering
	\subfigure{}
	\vspace{-4mm}
	\subfigure{
		\rotatebox[]{90}{\scriptsize{Input}}
		\hspace{1pt}
		\begin{minipage}{0.31\linewidth}
			\vspace{3pt}
			\centerline{\includegraphics[width=2.6cm,height=1.8cm]{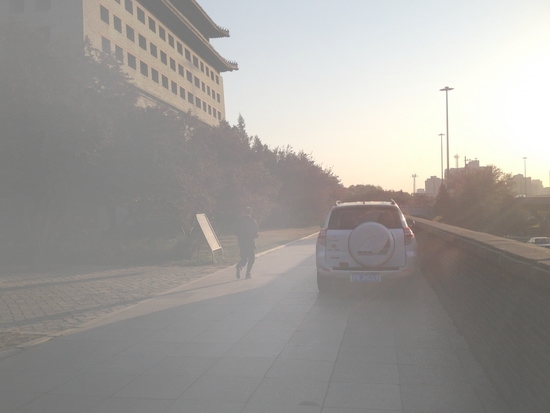}}
		\end{minipage}
		\begin{minipage}{0.31\linewidth}
			\vspace{3pt}
			\centerline{\includegraphics[width=2.6cm,height=1.8cm]{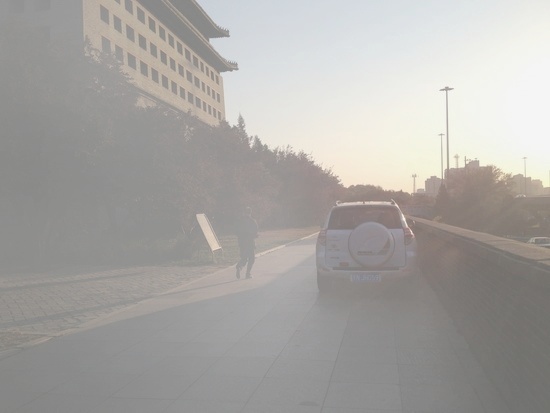}}
		\end{minipage}
		\begin{minipage}{0.31\linewidth}
			\vspace{3pt}
			\centerline{\includegraphics[width=2.6cm,height=1.8cm]{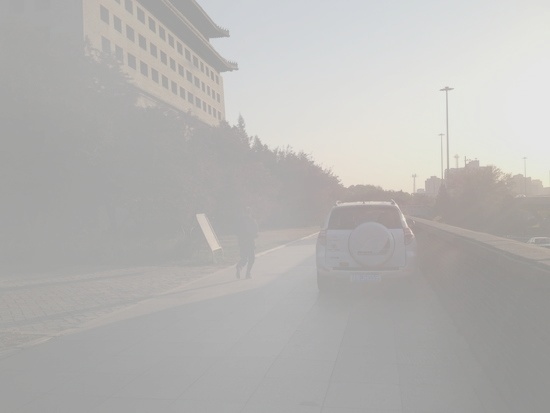}}
		\end{minipage}}
	\vspace{-4mm}
    \rotatebox[]{90}{\scriptsize{Dehazing Model 1}}
	\hspace{1pt}
	\subfigure{
		\begin{minipage}{0.31\linewidth}
			\vspace{3pt}
			\centerline{\includegraphics[width=2.6cm,height=2cm]{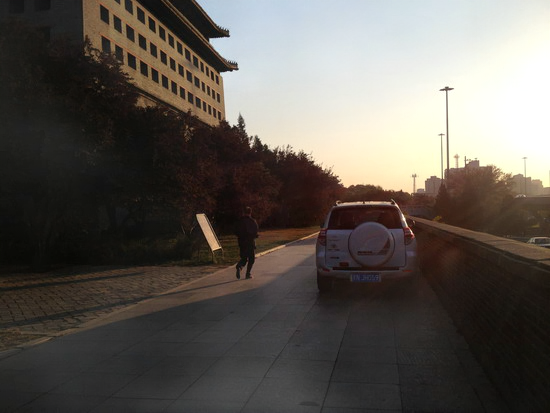}}
		\end{minipage}
		\begin{minipage}{0.31\linewidth}
			\vspace{3pt}
			\centerline{\includegraphics[width=2.6cm,height=2cm]{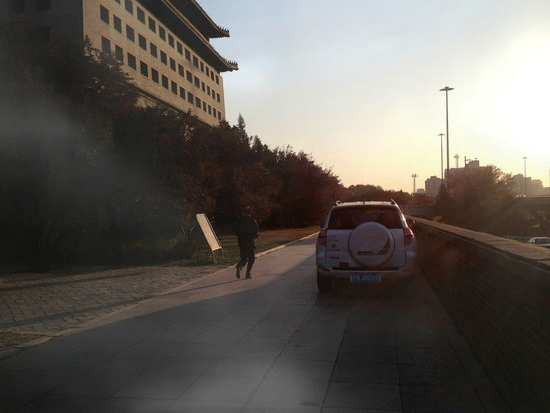}}
		\end{minipage}
		\begin{minipage}{0.31\linewidth}
			\vspace{3pt}
			\centerline{\includegraphics[width=2.6cm,height=2cm]{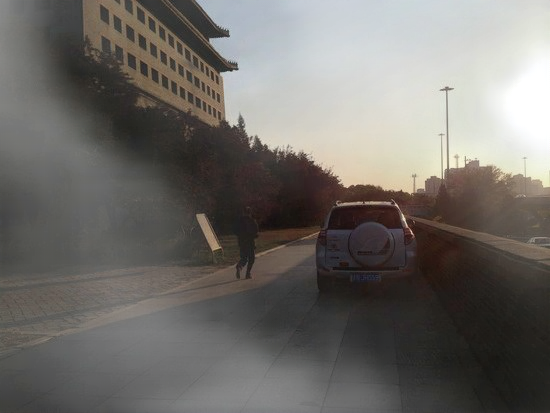}}
		\end{minipage}}
	\vspace{-4mm}
	\subfigure{
    \rotatebox[]{90}{\scriptsize{Dehazing Model 2}}
    \hspace{1pt}
		\begin{minipage}{0.31\linewidth}
			\vspace{3pt}
			\centerline{\includegraphics[width=2.6cm,height=2cm]{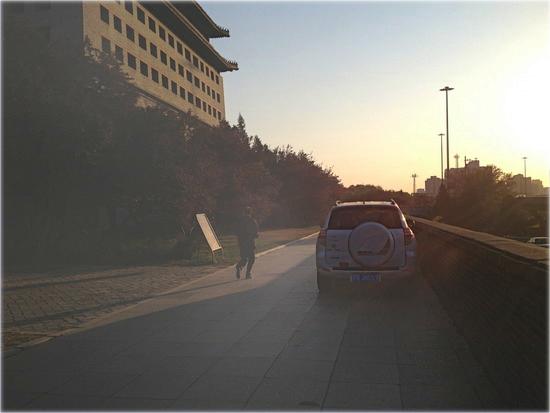}}
		\end{minipage}
		\begin{minipage}{0.31\linewidth}
			\vspace{3pt}
			\centerline{\includegraphics[width=2.6cm,height=2cm]{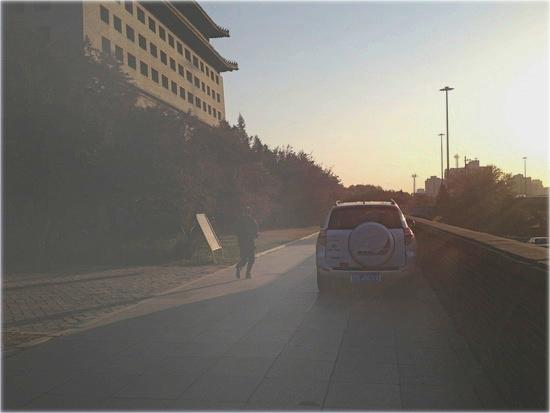}}
		\end{minipage}
		\begin{minipage}{0.31\linewidth}
			\vspace{3pt}
			\centerline{\includegraphics[width=2.6cm,height=2cm]{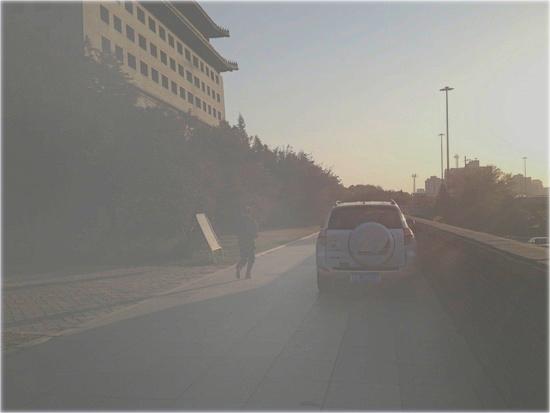}}
		\end{minipage}}
	\caption{The results of two dehazing methods on the same image with different hazy densities. The first row shows the input hazy images with three different densities. The second and third rows show the corresponding dehazed images by current two representative methods (denoted as ``Dehazing Model 1'' and ``Dehazing Model 2''), respectively. It shows that such methods are lack of robustness to cope with hazy images in different scenarios. }
	\label{visual_problem}
\end{figure}


Great efforts have been made in the past few years and end-to-end deep learning based dehazing methods has achieved great success in dealing with complex scenes~\cite{dong2020multi,li2017aod,hong2020distilling,shao2020domain}. However, when performing on scenarios with different haze density, these methods still cannot always obtain desirable dehzing performance witnessed by the inconsistency results as shown in Figure~\ref{visual_problem}. We can clear see that the same image with different hazy densities usually generate dehazed images with different qualities, by some current designed dehazing models. This phenomenon illustrates that such image dehazing models are not robust to some complex hazy scenarios, which is not what we expected for a good dehazing model. Unfortunately, this situation may usually happen in real world.  Consequently, how to improve the robustness of dehazing model becomes an important yet under-studied issue.



The inconsistency results shown in Figure~\ref{visual_problem} also inspire us to regularize the learning process by utilizing these multi-level hazy images with different densities, to improve the model robustness.
To achieve this goal, we creatively propose a Contrast-Assisted Reconstruction Loss (CARL) under the consistency-regularized framework for single image dehazing. It aims to train a more robust dehazing model to better deal with
various hazy scenarios, including but not limited to multi-level hazy images with different densities.


In the proposed CARL, we fully exploit the negative information to better facilitate the traditional positive-orient dehazing objective function. Specifically, we denote the restored dehazed image and its corresponding clear image (i.e. ground-truth) as anchor and positive example, respectively. The negative examples can be constructed not only from the original hazy image and its variants with different hazy densities, but also from the restored images generated by some other dehazing models.
The CARL enables the network prediction to be close to the clear image, and far away from the negative hazy images in the representation space. Specifically, pushing anchor point far away from various negative images seems to squeeze the anchor point to its positive example from different directions. Elaborately selecting the negative examples can help the CARL to improve the lower bound for approximating to its clear image under the regularization of various negative hazy images. Besides, more negative examples in the contrastive loss can usually contribute more performance improvement, which has been demonstrated in metric learning research field~\cite{khosla2020supervised}. Therefore, we also try to adopt more negative hazy images to improve the model capability to cope with various hazy scenarios.


To further improve the model robustness, we propose to train the dehazing network under the consistency-regularized framework on top of the CARL.
The success of consistency regularization lies in the assumption that the dehazing model should output very similar or even same dehazed images when fed the same hazy image with different densities. Such constraint meets the requirement of a good dehazing model to deal with hazy images in different hazy scenarios. Specifically, we implement this consistency constraint by the mean-teacher framework~\cite{tarvainen2017mean}. For each input hazy image, we also resort to previously constructed images with different hazy densities or some other informative negative examples.
Then, L1 loss is used to minimize the discrepancy among all the dehazed images which corresponds to one clear target. The consistency regularization significantly improves the model robustness and performance of the dehazing network, and it can be easily extended to any other dehazing network architectures.

Our main contributions are summarized as follows:
\begin{itemize}
  \item We make an early effort to consider dehazing robustness under variational haze density, which is a realistic while under-studied problem in the research filed of singe image dehazing.

  \item We propose a contrast-assisted reconstruction loss under the consistency-regularized framework for single image dehazing. This method can fully exploit various negative hazy images, to improve the dehzing model capability of dealing with various complex hazy scenarios.
  \item Extensive experimental results on  two synthetic and three real-world image dehazing datasets demonstrate that the proposed method significantly surpasses the state-of-the-art algorithms.
\end{itemize}
\begin{figure*}
  \centering
  \includegraphics[width=17.5cm]{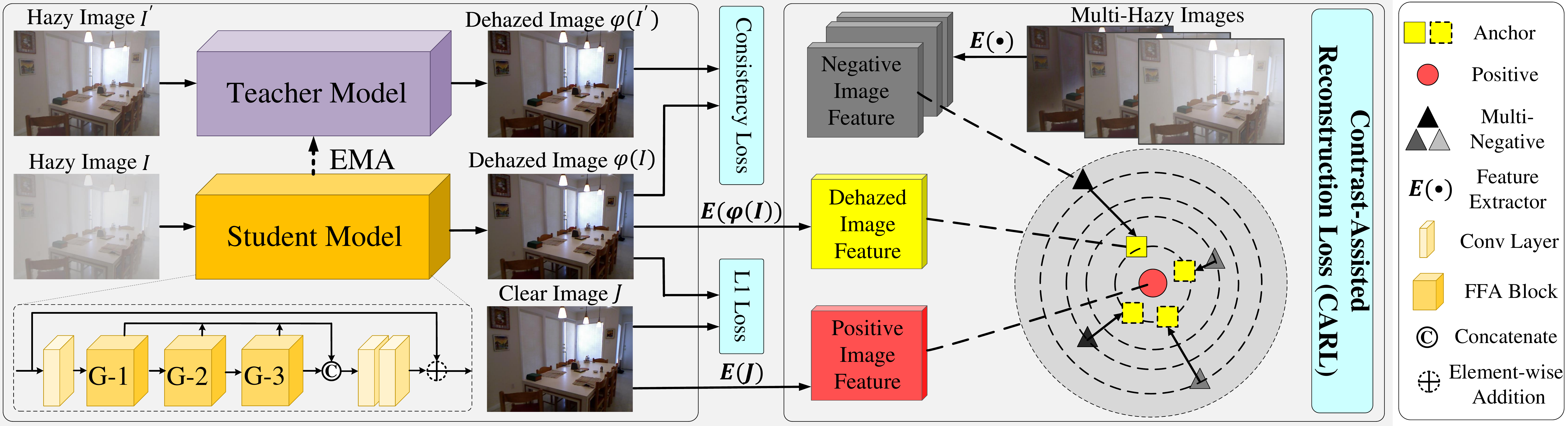}\\
   \vspace{-3mm}
   \caption{The framework of the proposed method for robust single image dehazing . It consists of the same dehazing network architecture for the student and teacher network, the Contrast-Assisted Reconstruction Loss (CARL) to fully exploit various negative hazy images to squeeze the dehazed image to its clean target from different directions, the consistency regularization framework to explicitly deal with multi-level hazy scenarios to further improve its robustness, and the traditional L1 reconstruction loss.}\label{framework}
\end{figure*}

\section{Related Work}
Single image dehazing aims to generate the hazy-free images from the hazy observations. We can roughly divide existing dehazing methods into categories: the physical-scattering-model dependent methods and the model-free methods.

\textbf{The physical-scattering-model dependent methods} try to recover the clear image through estimating the atmospheric light and transmission map by some specially designed priors or network architectures. For the prior-based image dehazing methods,  they usually remove the haze using different statistic image prior from empirical observations, such as the traditional dark channel prior (DCP)~\cite{he2010single}, non-local prior~\cite{berman2016non} and contrast maximization~\cite{tan2008visibility}. Although these methods have achieved a series of successes, the priors can not handle all the cases in the unconstraint wild environment. For instance, the classical DCP~\cite{he2010single} can not well dehaze the sky regions in the hazy image, since it does not satisfy the prior assumption.


Recently, as the prevailing success of deep learning in image processing tasks, many deep dehazing methods depending on the atmosphere scattering model have been proposed.
~\cite{zhang2018densely} directly embedded the physical model into the dehazing network by a densely connected encoder-decoder structure.
~\cite{ren2016single} proposed a coarse-to-fine multi-scale convolutional neural network to estimate the transmission map.~\cite{li2017aod} reformulated the original atmospheric scattering model and jointly estimated the global atmospheric light and the transmission map.
However, such physical-scattering-model based learning methods may produce accumulative error and degrade the dehazing results, since the inaccurate estimation or some estimation bias on the transmission map and global atmospheric light results in larger reconstruction error between the restored images and the clear ones. Besides, it is difficult or even unable to collect the ground-truth about transmission map and global atmospheric light in the real-world scenarios.

\textbf{Model-free deep dehazing methods} try to directly learn the map between the hazy input and clean result without using atmospheric scattering model. Most of such methods focus on strengthening the dehazing network. For instance, ~\cite{liu2019dual} designed a dual residual neural network architecture to explore the potential of paired operations for image restoration tasks. ~\cite{qu2019enhanced} proposed a pixel-to-pixel dehazing network to obtain perceptual pleasing results. ~\cite{qin2020ffa} proposed an attention fusion mechanism to enhance the flexibility of the network to deal with different types of information. ~\cite{chen2019gated} proposed an end-to-end gated context aggregation network to restore the hazy-free image. These methods only minimize the reconstruction loss between the restored dehazed image and its clean target, without any regularization on images or features.

Recently, there also appeared some methods, which adopted distance metric regularization to further improve the reconstruction loss. ~\cite{wu2021contrastive} proposed the divided-contrast regularization for single image dehazing. Our method also falls into this category, but it is very different from them. We specifically propose the CARL under the consistency regularization framework for single image dehazing.
It can not only fully exploit existing negative hazy examples to make the dehazing model generate more natural restored images, but also can further improve the model robustness to deal with various complex hazy scenarios.



\section{Proposed Method}
The proposed method makes great efforts to deal with various hazy scenarios from the following two aspects: 1) The CARL improves the traditional positive-orient dehazing objective function, by fully exploiting the various negative information to squeeze the dehazed image to its clean target from different directions; 2) The consistency regularization aims to further improve the model robustness by explicitly utilizing the constraint outputs of multi-level hazy images. In the following, we describe the algorithm in detail as illustrated in Figure~\ref{framework}.

\subsection{Dehazing Network Architecture}
In this paper, we adopt the previously proposed FFA-Net~\cite{qin2020ffa} as our backbone network architecture. As shown in Figure~\ref{framework}, the student and teacher network share the same network architecture (FFA-Net), which includes the following components: the shallow feature extraction part, several group attention architecture (Denoted as G-x), the feature concatenation module, the reconstruction part and global residual skip connection.

Specifically, the input hazy image is first processed by one convolution layer with kernel size of $3\times 3$, to extract shallow features. Then, it passes through three group architectures, where each of them consists of 19 basic blocks architecture (BBA), and each BBA is constructed by skip connection and the feature attention module~\cite{qin2020ffa}. Followed by the concatenation of the outputs from the three group architectures in channel-wise orientation, the features pass through two convolution layers combined with the global skip connection. Finally, the hazy-free image is obtained.

\subsection{Contrast-Assisted Reconstruction Loss}
The contrastive learning method has achieved a series of successes in representation learning, it aims to learn discriminative feature representations by pulling ``positive'' pairs close, while pushing ``negative'' pairs far apart. Inspired by this, we propose the ``Contrast-Assisted'' Reconstruction Loss (CARL) to improve the traditional positive-orient dehazing methods, by fully exploiting various negative information to squeeze the dehazed image to its clean target from different directions.


To imitate the traditional contrastive learning, there are two aspects we need to consider: one is how to construct the positive and negative training examples, the other is how to apply the CARL in the dehazing framework~\cite{wu2021contrastive}. As we known, elaborately constructing efficient positive and negative training examples is very crucial to better optimize the CARL. For the image dehazing task, obviously the positive pair is the dehazed image and its corresponding clear one, which can be denoted as anchor point and positive example. Our final goal is just to minimize the discrepancy between them. Meanwhile, pushing anchor point far away from several negative examples is to squeeze the anchor point to positive example from different directions, as illustrated in Figure~\ref{framework}. Therefore, we generate negative examples from several aspects, which includes the original hazy image, multi-level hazy images with different densities, some relatively low-quality dehazed images by previous model, and some other variants of the input hazy images. For the latent space to apply the CARL, we adopt the commonly used intermediate feature from the fixed pre-trained model ``E'' to works as the feature extractor, e.g. VGG-19~\cite{simonyan2014very}, which was used in~\cite{wu2021contrastive,johnson2016perceptual}.

Denote the input hazy image as $I$, its corresponding dehazed image as $\phi(I)$ which is generated by the dehazing network $\phi$, and the ground-truth hazy-free image as $J$. The selected negative images corresponding to $\phi(I)$ denote as $I^-_i, i\in \{1,\cdots,K\}$, $K$ is the number of negative examples used in the CARL. We define the features extracted by the fixed pre-trained VGG model as $E(J)$, $E(\phi(I))$ and $E(I^-_i)$ for the positive, anchor and negative examples, respectively. Then, the $m$-th CARL function $\mathcal{L}^m_{CARL}$ can be formulated as:

\begin{equation}\label{EQ:CARLm}
-\log\frac{\textmd{e}^{-|E_m(\phi(I))-E_m(J)|/\tau}}{\textmd{e}^{-|E_m(\phi(I))-E_m(J)|/\tau}+\sum_{i=1}^K \textmd{e}^{-|E_m(\phi(I))-E_m(I^-_i)|/\tau}}.
\end{equation}
In Eq.~\ref{EQ:CARLm}, ``$\textmd{e}$'' denotes the exponential operation, $E_m, m=\{1,2,\cdots, M\}$, extracts the $m$-th hidden features from the fixed pre-trained model VGG~\cite{simonyan2014very}. $|\cdot|$ demotes the $L1$ distance, which usually achieves better performance compared to $L2$ distance for image dehazing task. $\tau > 0$ is the temperature parameter that controls the sharpness of the output. Therefore, the final CARL can be expressed as follows,
\begin{equation}\label{EQ:CARL}
\mathcal{L}_{CARL} = \sum_{m=1}^{M}\omega_m \mathcal{L}_{CARL}^m,
\end{equation}
where $\omega_m$ is the weight coefficient for the $m$-th CARL using the intermediate feature generated by the fixed VGG model.

Note that, in Eq.~\ref{EQ:CARLm}, the positive point $E_m(J)$ and all the negative points $E_m(I^-_i)$ are constant values, minimizing $\mathcal{L}_{CARL}$ can optimize the parameters of the dehazing network $\phi$ through the dehazed image features $E_m(\phi(I))$.
Related to our CARL, perceptual loss~\cite{johnson2016perceptual} minimizes the visual difference between the prediction and ground truth by using multi-layer features extracted from the fixed pre-trained deep model. On top of this, one divided-contrastive learning method~\cite{wu2021contrastive} adopted the original hazy image as negative image to regularize the solution space. Different from above methods, the proposed CARL method aims to minimize the reconstruction error between the prediction and its corresponding ground truth, as well as pushing the prediction far away from various negative hazy examples, which acts as a way to squeeze prediction to its ground truth from different directions in the constraint learning space. Thus, it enables the dehazing model to deal with various complex hazy scenarios.

The main difference between the traditional contrastive learning and our proposed CARL is that:
The traditional contrastive learning aims to learn discriminative feature representations to distinguish instances from different classes or identities, which cares about both the inter-class discrepancy and intra-class compactness. However, for such image dehazing task, we just consider the reconstruction error between the dehazed image and its corresponding clean target. Therefore, our final goal of CARL is to better optimize the reconstruction loss with the help of various negative hazy examples in the contrastive manner. This is also the reason why we call the proposed method ``Contrast-Assisted Reconstruction loss''.


\subsection{The Consistency-Regularized Framework}
We propose the consistency regularization based on the assumption that a good dehazing model should output very similar or even same dehazed images when fed the same hazy image with different densities. Such explicit constraint further improves the model robustness to deal with multi-level hazy images. This learning paradigm is implemented by training a student neural network $\phi_s(\cdot)$ and a teacher neural network $\phi_t(\cdot)$, which share the same network architecture,  but are parameterized by $\theta_s$ and $\theta_t$ respectively.

Specifically, we first construct hazy images with different densities using the physical-scattering model. Usually, the synthetic dehazing datasets themselves contain several hazy images with different densities corresponding to one clean image. For the real-world dehazing dataset, we generate different hazy images with the help of the transmission map and atmospheric light given for other dataset, and some relatively poor-quality dehazed image obtained by previous dehazing model, such as DCP~\cite{he2010single}.

Given two different hazy images denoted as  $I$ and $I'$, which correspond to the same clear image $J$, they are fed into the student and teacher network respectively. These two hazy images are processed by dehazing network $\phi_s(\cdot)$ and $\phi_t(\cdot)$ to obtain the dehazed images $\phi_s(I)$ and $\phi_t(I')$. The consistency regularization $\mathcal{L}_{CR}$ can be expressed as follows,
\begin{equation}\label{Eq:Consistency}
  \mathcal{L}_{CR}=|\phi_s(I)-\phi_t(I')|.
\end{equation}
In Eq.~\ref{Eq:Consistency}, we use the $L_1$ loss to implement the consistency regularization. Minimizing the loss function $\mathcal{L}_{CR}$ can directly optimize parameters of the student network $\phi_s(\cdot)$, while the parameters of the teacher network is updated by the exponential moving average (EMA) techniques, which is based on the previous teacher network and current student network parameters. Unlike previous teacher network for image dehazing~\cite{hong2020distilling}, we do not have a predefined high quality model as the fixed teacher model, we build it from past iteration of the student network $\phi_s(\cdot)$. The updating rule of ``EMA'' is $\theta_t\leftarrow \lambda\theta_t + (1-\lambda)\theta_s$, and $\lambda$ is a smoothing hyper-parameter to control the model updating strategy~\cite{tarvainen2017mean}.

Therefore, such a consistency regularization keeps hazy image with different densities or under different scenarios, have the same haze-free output, which greatly improves the robustness of the image dehazing model.

\subsection{The Overall Loss Function}
Apart from above introduced CARL and the consistency regularization, we also adopt the traditional reconstruction loss $\mathcal{L}_{1}$ between the prediction $\phi(I)$ and its corresponding ground truth $J$ in the data field. It can be implemented by the $L1$ loss as follows,
\begin{equation}\label{Eq:Reconstruction}
  \mathcal{L}_{1}=|\phi_s(I)-J|.
\end{equation}

Therefore, the overall loss function $\mathcal{L}$ can be expressed as,
\begin{equation}\label{Eq:OverallLoss}
  \mathcal{L} = \mathcal{L}_{1} + \lambda_1 \mathcal{L}_{CR} + \lambda_2\mathcal{L}_{CARL},
\end{equation}
where $\lambda_1$ and $\lambda_2$ are two hyper-parameters to balance above three terms in the overall loss function.


\section{Experiments}
\subsection{Experiment Setup}
\textbf{Datasets and Metrics.}
To comprehensively evaluate the proposed method, we conduct extensive experiments on two representative synthetic datasets and three challenging real-world datasets. The RESIDE dataset \cite{li2018benchmarking} is a widely used synthetic dataset, which consists of two widely used subsets, i.e., Indoor Training Set (ITS), Outdoor Training Set (OTS). ITS and OTS are used as the training dataset, and they have corresponding testing dataset, namely, Synthetic Objective Testing Set (SOTS), which contains 500 indoor hazy images (SOTS-Indoor) and 500 outdoor hazy ones (SOTS-Outdoor). We also evaluate the proposed model on the following real-world datasets: NTIRE 2018 image dehazing indoor dataset (referred to as I-Haze) \cite{ancuti2018ihaze}, NTIRE 2018 image dehazing outdoor dataset (O-Haze) \cite{ancuti2018ohaze}, and NTIRE 2019 dense image dehazing dataset (Dense-Haze) \cite{ancuti2019dense}. We conduct objective and subjective measurement respectively. For the objective measurement, we adopt Peak Signal to Noise Ratio (PSNR) and the Structural Similarity index (SSIM) as evaluation metrics, which are widely used to evaluate image quality for the image dehazing task. For the subjective measurement, we evaluate the performance by comparing the visual effect of dehazed images.

\textbf{Implementation Details.}
We implement the proposed method based on PyTorch with NVIDIA GTX 2080Ti GPUs. In the entire training process, we randomly crop $240 \times 240$ image patches as input and adopt Adam optimizer with the default exponential decay rate $0.9$ for optimization. The learning rate is initially set to $1 \times 10^{-4}$ and is adjusted using the cosine annealing strategy \cite{he2019bag}. We follow \cite{wu2021contrastive} to select the features of the 1st, 3rd, 5th, 9th and 13th layers from the fixed pre-trained VGG-19 \cite{simonyan2014very} to calculate the L1 distance in Eq.(\ref{EQ:CARL}), and their corresponding weight factors $\omega_m$ are set as $\frac{1}{32}$, $\frac{1}{16}$, $\frac{1}{8}$, $\frac{1}{4}$ and $1$, respectively. The number of negative examples used in Eq.~\ref{EQ:CARLm} is set to $K=5$, and parameter $\tau=0.5$. The hyper-parameters $\lambda_1$ and $\lambda_2$ in Eq.~\ref{Eq:OverallLoss} is set to 1.0 and 10.0 respectively.

\begin{table}[]
	\scriptsize
	\centering
    \caption{Quantitative comparisons with SOTA methods on SOTS-Indoor and SOTS-Outdoor synthetic datasets.}
	\vspace{-2mm}
	\setlength\tabcolsep{3.5pt}
	\scalebox{1}[1]{
		\begin{tabular}{cccccc}
			\toprule
			\multicolumn{1}{c}{\multirow{3}{*}{Method}}
			&\multicolumn{1}{c}{\multirow{3}{*}{Reference}}
			& \multicolumn{2}{c}{SOTS-Indoor}
			& \multicolumn{2}{c}{SOTS-Outdoor}
			\\
			\cmidrule(l){3-6}
			& & PSNR & SSIM  & PSNR & SSIM     \\
			\midrule
			DCP \cite{he2010single}              & TPAMI  & 16.62 & 0.8179  & 19.13 & 0.8148    \\
			MSCNN \cite{ren2016single}            & ECCV   & 17.57 & 0.8102  & 20.73 & 0.8187    \\
			DehazeNet \cite{cai2016dehazenet}        & TIP    & 21.14 & 0.8472  & 22.46 & 0.8514    \\
			AOD-Net \cite{li2017aod}          & ICCV   & 19.06 & 0.8504  & 20.29 & 0.8765    \\
			DCPDN \cite{zhang2018densely}            & CVPR   & 19.00 & 0.8400  & 19.71 & 0.8300    \\
			GFN \cite{ren2018gated}              & CVPR   & 22.30 & 0.8800  & 21.55 & 0.8444    \\
			EPDN \cite{qu2019enhanced}             & CVPR   & 25.06 & 0.9232  & 22.57 & 0.8630    \\
			DuRN-US \cite{liu2019dual}          & CVPR   & 32.12 & 0.9800  & 19.41 & 0.8100    \\
			GridDehazeNet \cite{liu2019griddehazenet}    & ICCV   & 32.16 & 0.9836  & 30.86 & 0.9819    \\
			KDDN \cite{hong2020distilling}             & CVPR   & 34.72 & 0.9845  & --    & --        \\
			DA \cite{shao2020domain}               & CVPR   & 25.30 & 0.9420  & 26.44 & 0.9597    \\
			MSBDN \cite{dong2020multi}            & CVPR   & 32.00 & 0.9860  & 30.77 & 0.9550    \\
			FFA-Net \cite{qin2020ffa}          & AAAI   & 36.39 & 0.9886  & \underline{32.09} & 0.9801    \\
			FD-GAN \cite{dong2020fd}           & AAAI   & 23.15 & 0.9207  & --    & --        \\
			DRN \cite{li2020deep}              & TIP    & 32.41 & 0.9850  & 31.17 & \underline{0.9830}    \\
			AECR-Net \cite{wu2021contrastive}         & CVPR   & 37.17 & \underline{0.9901}  & --    & --        \\
			DIDH \cite{yoon2021towards}             & AAAI   & \underline{38.91} & 0.9800  & 30.40 & 0.9400    \\
			\midrule
			\multicolumn{1}{c}{\textbf{Ours}}       & --  & \textbf{41.92} & \textbf{0.9954}  &\textbf{33.26} & \textbf{0.9849}  \\
			\bottomrule
	\end{tabular}}
	
	\label{SOTS_table}
\end{table}

\subsection{Comparisons with State-of-the-art Methods}
\begin{figure*}[]
	\centering
	\begin{minipage}{0.1\linewidth}
		\vspace{3pt}
		\centerline{\includegraphics[width=1.8cm,height=2cm]{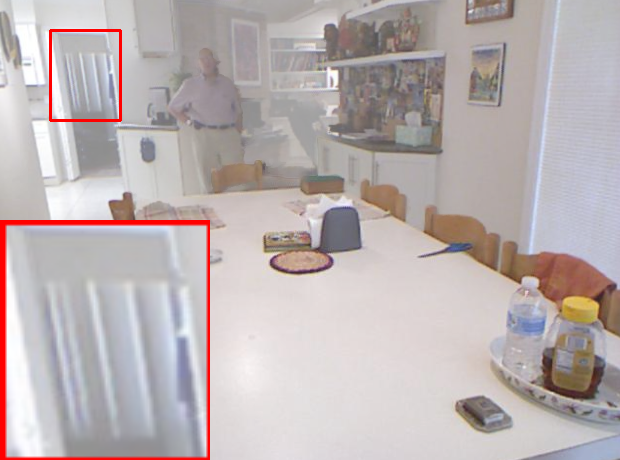}}
		\centerline{14.00 / 0.87}
		\vspace{3pt}
		\centerline{\includegraphics[width=1.8cm,height=2cm]{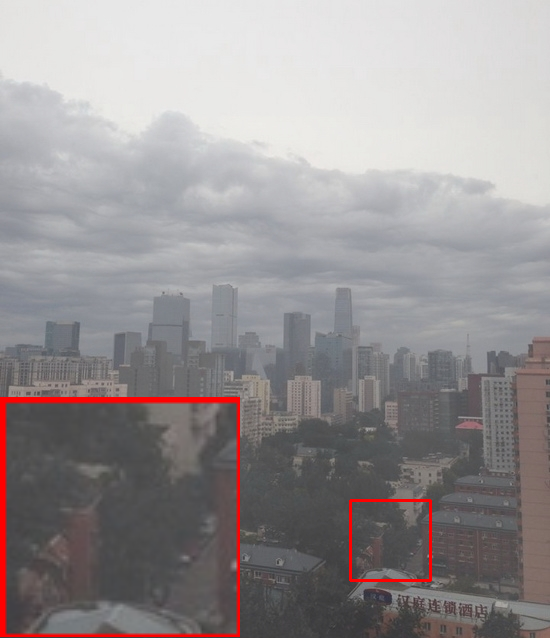}}
		\centerline{15.89 / 0.80}
		\vspace{3pt}
		\centerline{Input}
	\end{minipage}
	\begin{minipage}{0.1\linewidth}
		\vspace{3pt}
		\centerline{\includegraphics[width=1.8cm,height=2cm]{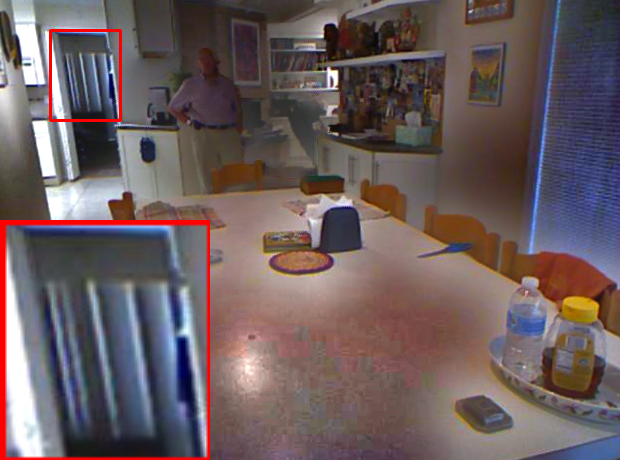}}
		\centerline{11.74 / 0.79}
		\vspace{3pt}
		\centerline{\includegraphics[width=1.8cm,height=2cm]{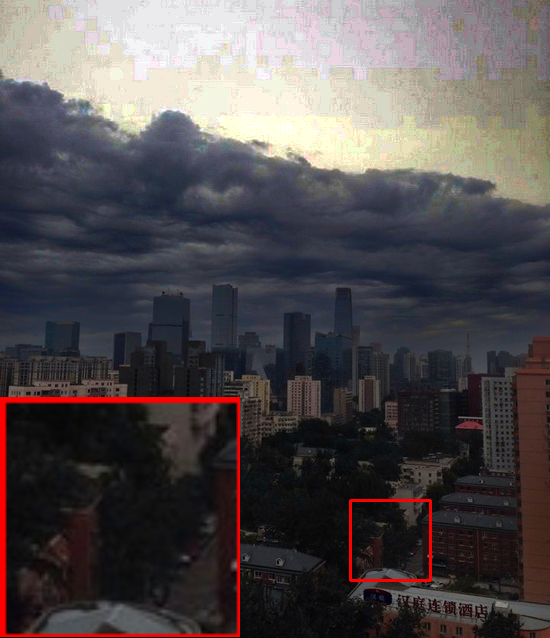}}
		\centerline{13.09 / 0.79}
		\vspace{3pt}
		\centerline{DCP}
	\end{minipage}
	\begin{minipage}{0.1\linewidth}
		\vspace{3pt}
		\centerline{\includegraphics[width=1.8cm,height=2cm]{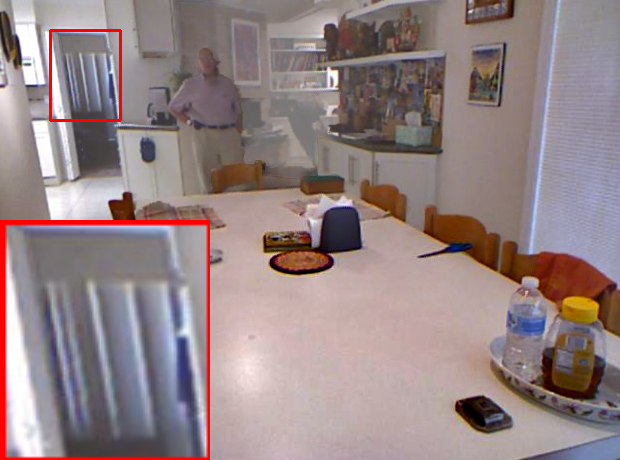}}
		\centerline{17.17 / 0.89}
		\vspace{3pt}
		\centerline{\includegraphics[width=1.8cm,height=2cm]{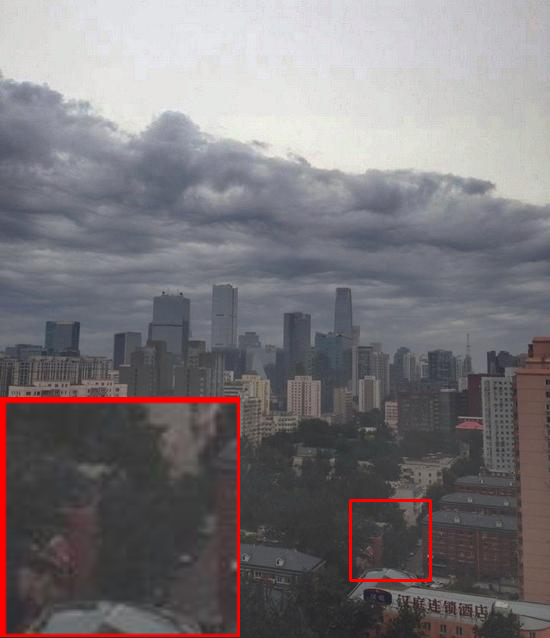}}
		\centerline{18.16 / 0.82}
		\vspace{3pt}
		\centerline{MSCNN}
	\end{minipage}
	\begin{minipage}{0.1\linewidth}
		\vspace{3pt}
		\centerline{\includegraphics[width=1.8cm,height=2cm]{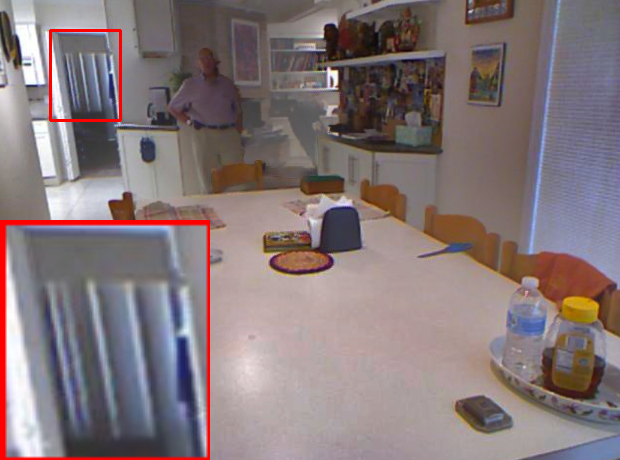}}
		\centerline{17.07 / 0.91}
		\vspace{3pt}
		\centerline{\includegraphics[width=1.8cm,height=2cm]{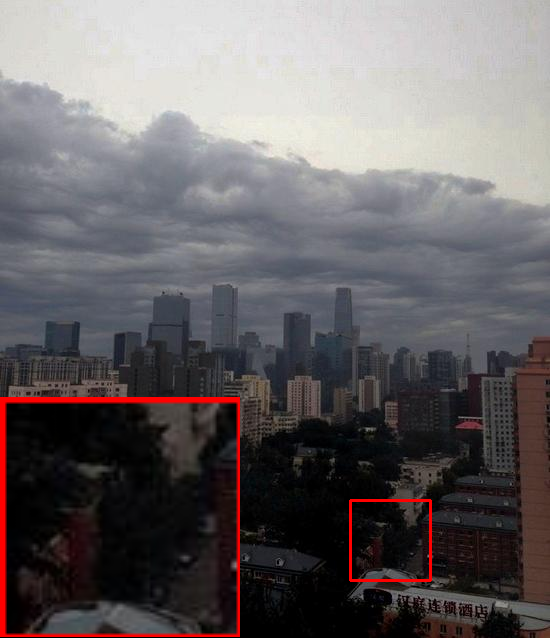}}
		\centerline{21.26 / 0.85}
		\vspace{3pt}
		\centerline{DehazeNet}
	\end{minipage}
	\begin{minipage}{0.1\linewidth}
		\vspace{3pt}
		\centerline{\includegraphics[width=1.8cm,height=2cm]{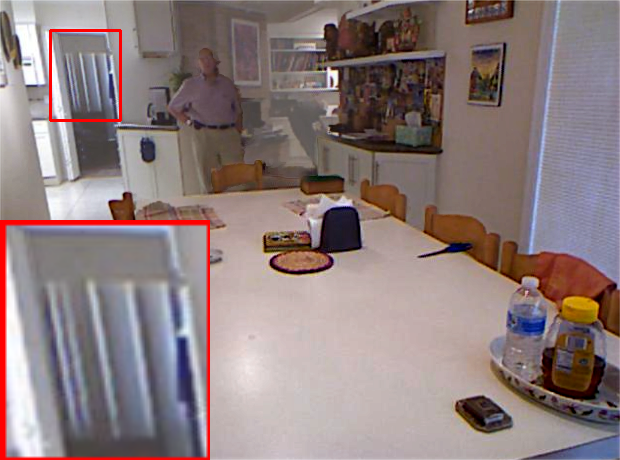}}
		\centerline{17.60 / 0.89}
		\vspace{3pt}
		\centerline{\includegraphics[width=1.8cm,height=2cm]{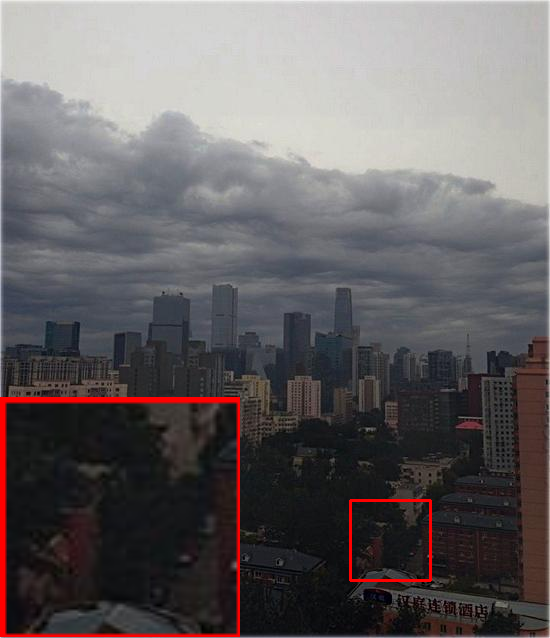}}
		\centerline{18.81 / 0.90}
		\vspace{3pt}
		\centerline{AOD-Net}
	\end{minipage}
	\begin{minipage}{0.1\linewidth}
		\vspace{3pt}
		\centerline{\includegraphics[width=1.8cm,height=2cm]{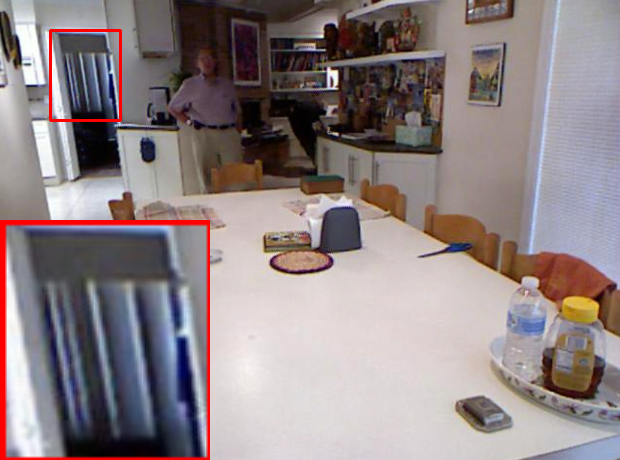}}
		\centerline{30.72 / 0.98}
		\vspace{3pt}
		\centerline{\includegraphics[width=1.8cm,height=2cm]{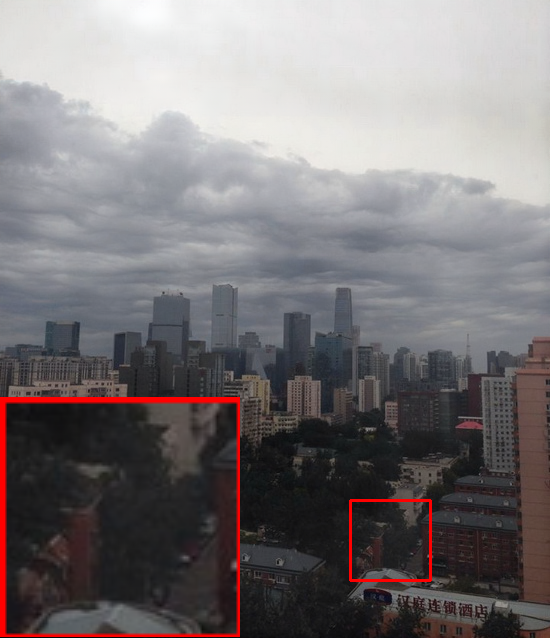}}
		\centerline{28.85 / 0.97}
		\vspace{3pt}
		\centerline{GridDehazeNet}
	\end{minipage}
	\begin{minipage}{0.1\linewidth}
		\vspace{3pt}
		\centerline{\includegraphics[width=1.8cm,height=2cm]{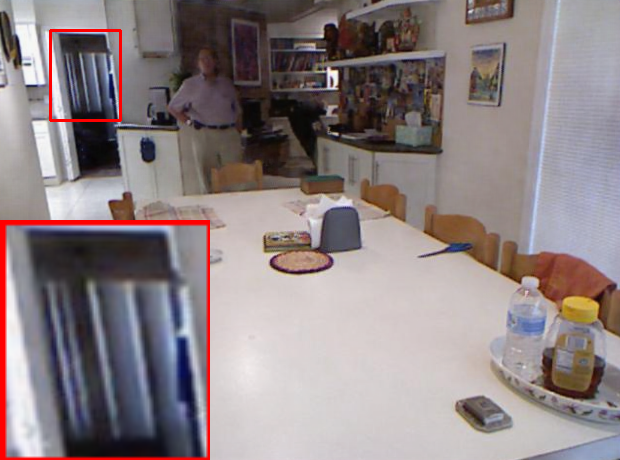}}
		\centerline{33.05 / 0.98}
		\vspace{3pt}
		\centerline{\includegraphics[width=1.8cm,height=2cm]{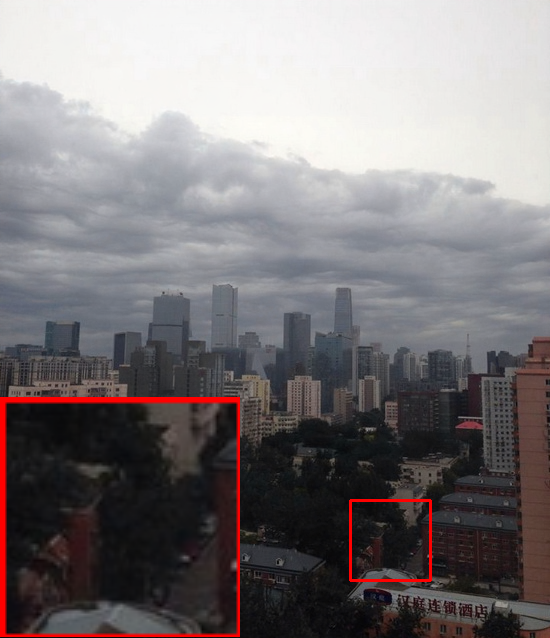}}
		\centerline{33.22 / $0.98_4$}
		\vspace{3pt}
		\centerline{FFA-Net}
	\end{minipage}
	\begin{minipage}{0.1\linewidth}
		\vspace{3pt}
		\centerline{\includegraphics[width=1.8cm,height=2cm]{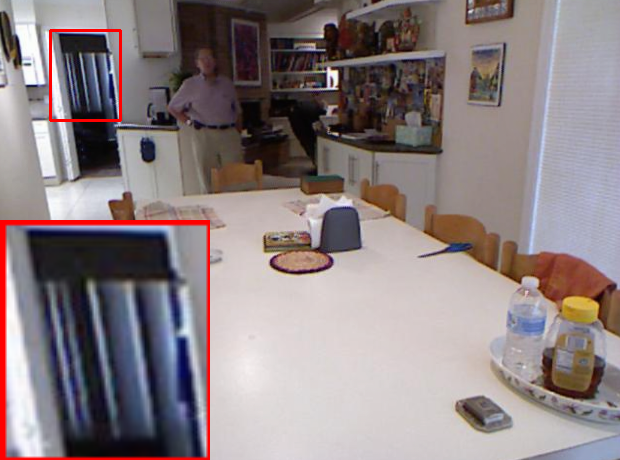}}
		\centerline{39.71 / $0.99_5$}
		\vspace{3pt}
		\centerline{\includegraphics[width=1.8cm,height=2cm]{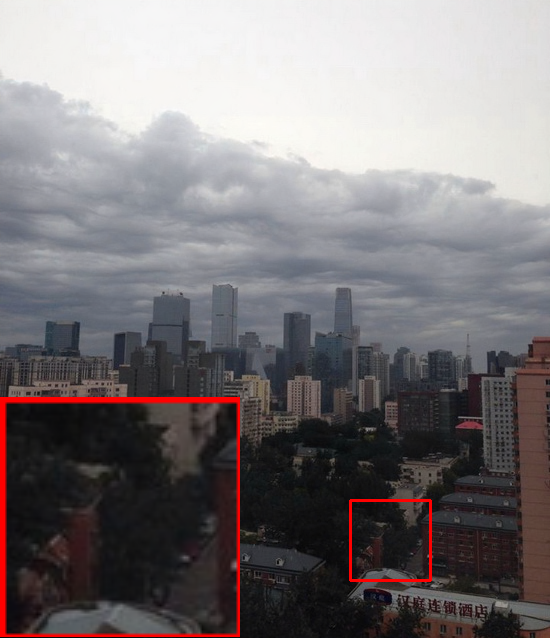}}
		\centerline{36.09 / $0.98_6$}
		\vspace{3pt}
		\centerline{Ours}
	\end{minipage}
	\begin{minipage}{0.1\linewidth}
		\vspace{3pt}
		\centerline{\includegraphics[width=1.8cm,height=2cm]{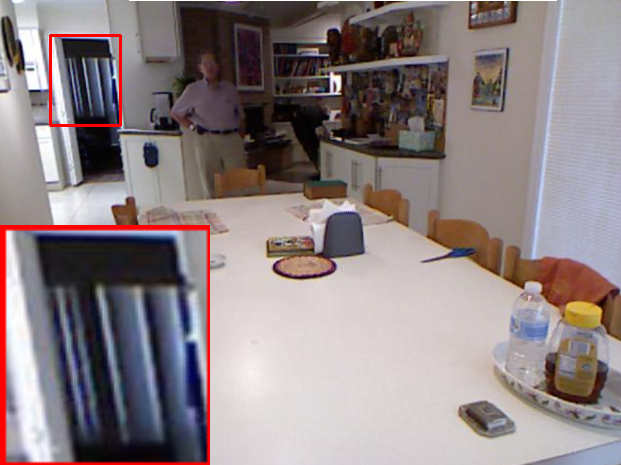}}
		\centerline{$\infty$ / 1}
		\vspace{3pt}
		\centerline{\includegraphics[width=1.8cm,height=2cm]{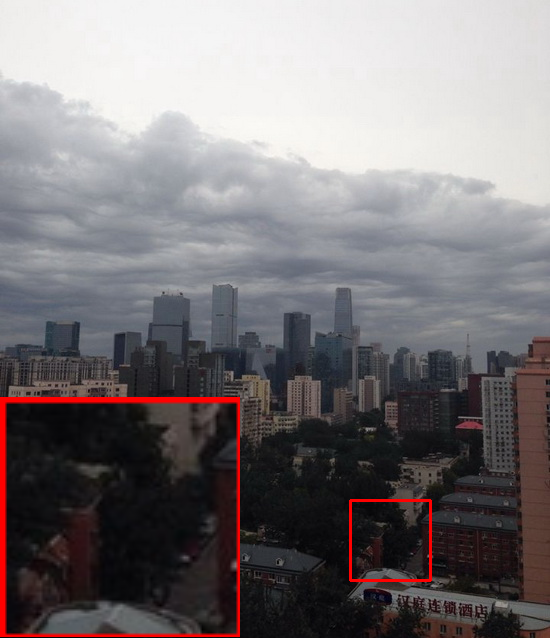}}
		\centerline{$\infty$ / 1}
		\vspace{3pt}
		\centerline{Ground Truth}
	\end{minipage}
	\caption{Visual comparisons on two hazy images with different methods. The first row: the dehazed results from SOTS-Indoor dataset. The second row: the dehazed results from SOTS-Outdoor dataset. The numbers under each image represent the PSNR and SSIM values.}
	\label{VisualImage}
	\vspace{-2mm}
\end{figure*}

\textbf{Results on Synthetic Datasets.}
We follow the settings of ~\cite{qin2020ffa} to evaluate our proposed method on two representative synthetic datasets, and compare it with seventeen state-of-the-art (SOTA) methods.
The results are shown in Table \ref{SOTS_table}. We can see that our proposed method achieves the best performance on both SOTS-Indoor and SOTS-Outdoor datasets. On SOTS-Indoor testing dataset, our proposed method achieves 41.92dB PSNR and 0.9954 SSIM, surpassing the second-best 3.01dB PSNR and 0.0053 SSIM, respectively. On the SOTS-Outdoor testing dataset, our proposed method achieves the gain with 1.17dB PSNR and 0.0019 SSIM, compared with the second-best method. In Figure~\ref{VisualImage}, we show the visual effects of representative methods of dehazed images for subjective comparison. We can observe that DCP suffers from the color distortion, where the images are darker and unrealistic. MSCNN, DehazeNet and AOD-Net cannot remove haze completely, and there are still a lot of haze residues. Although GridDehazeNet and FFA-Net can achieve better dehazed effect, there is still a gap between their results and the ground truth. In contrast, the images generated by our proposed method are closer to ground truth and more natural, which are also verified by the PSNR and SSIM.

\begin{table}[]
	\scriptsize
	\centering
    \caption{Quantitative comparisons with SOTA methods on I-Haze, O-Haze and Dense-Haze real-world datasets.}
	\vspace{-2mm}
	\setlength\tabcolsep{3.5pt}
	\scalebox{0.93}[1]{
		\begin{tabular}{ccccccc}
			\toprule
			\multicolumn{1}{c}{\multirow{3}{*}{Method}}
			
			& \multicolumn{2}{c}{I-Haze}
			& \multicolumn{2}{c}{O-Haze}
			& \multicolumn{2}{c}{Dense-Haze}
			\\
			\cmidrule(l){2-7}
			 & PSNR & SSIM  & PSNR & SSIM  & PSNR & SSIM  \\
			\midrule
			DCP \cite{he2010single}          & 14.43 & 0.7520    & 16.78 & 0.6530  & 10.06 & 0.3856  \\
			MSCNN \cite{ren2016single}        & 15.22 & 0.7550    & 17.56 & 0.6500  & 11.57 & 0.3959   \\
			DehazeNet \cite{cai2016dehazenet}  & 14.31 & 0.7220    & 16.29 & 0.6400  & 13.84 & 0.4252     \\
			AOD-Net \cite{li2017aod}            & 13.98 & 0.7320    & 15.03 & 0.5390  & 13.14 & 0.4144  \\
			GridDehazeNet \cite{liu2019griddehazenet}  & 16.62 & 0.7870    & 18.92 & 0.6720  & 13.31 & 0.3681   \\
			KDDN \cite{hong2020distilling}                & --    & --   & \underline{25.46} & \underline{0.7800}  & 14.28 & 0.4074   \\
			FFA-Net \cite{qin2020ffa}                & --    & --        & -- & -- & 14.39 & 0.4524   \\
			MSBDN \cite{dong2020multi}                & \underline{23.93}    & \textbf{0.8910}        & 24.36 & 0.7490 &15.37 &\underline{0.4858}   \\
			IDRLP \cite{ju2021idrlp}                & 17.36    & 0.7896        & 16.95 & 0.6990 & -- & --   \\
			AECR-Net \cite{wu2021contrastive}                & --    & --        & -- & -- & \textbf{15.80} & 0.4660   \\
			\midrule
			\multicolumn{1}{c}{\textbf{Ours}}         & \textbf{25.43} & \underline{0.8807}  &\textbf{25.83} & \textbf{0.8078}  &\underline{15.47} & \textbf{0.5482}  \\
			\bottomrule
	\end{tabular}}
	\label{IO_table}
\end{table}
\textbf{Results on Real-world Datasets.}
We also evaluate our proposed method on three challenging real-world datasets. As shown in Table \ref{IO_table}, our proposed method outperforms most state-of-the-art methods, and we have obtained 25.43dB PSNR and 0.8807 SSIM on I-Haze dataset, 25.83dB PSNR and 0.8078 SSIM on O-Haze dataset, 14.47db PSNR and 54.82 SSIM on Dense-Haze dataset.
On I-Haze and Dense-Haze datasets, although our proposed method is slightly lower than the state-of-the-art methods in SSIM and PSNR, but higher in terms of PSNR and SSIM, respectively. In general, our proposed method can still maintain advanced results on real-world datasets, and some visual comparison presented in Figure~\ref{VisualImage} also verifies it.

\begin{table}[]
	\scriptsize
	\centering
	\caption{Ablation study of the proposed method with different components on SOTS-Indoor and Dense-Haze datasets.}
	\vspace{-2mm}
	\setlength\tabcolsep{8pt}
	\scalebox{1}[1]{
		\begin{tabular}{ccccc}
			\toprule
			\multicolumn{1}{c}{\multirow{3}{*}{Method}}
			& \multicolumn{2}{c}{SOTS-Indoor}
			& \multicolumn{2}{c}{Dense-Haze}
			\\
			\cmidrule(l){2-5}
			 & PSNR & SSIM  & PSNR & SSIM   \\
			\midrule
			$\mathcal{L}_1$             & 36.39 & 0.9886    & 14.39 & 0.4524   \\
            $\mathcal{L}_1$+$\mathcal{L}_{DivC}$~\cite{wu2021contrastive}      & 37.54 & 0.9915    & 14.82 & 0.5354   \\
			$\mathcal{L}_1$+$\mathcal{L}_{CARL}$         & 39.56 & 0.9939    & 15.30 & 0.5438       \\
			$\mathcal{L}_1$+$\mathcal{L}_{CARL}$ + $\mathcal{L}_{CR}$          &\textbf{41.92} & \textbf{0.9954} & \textbf{15.47} & \textbf{0.5482}  \\

		\bottomrule
	\end{tabular}}
	\label{Analysis}
\end{table}

\begin{table}[h]
	\scriptsize
	\centering
	\caption{Parameter sensitivity analysis on $\lambda_2$ with $\lambda_1=1.0$ on SOTS-Indoor dataset.}
	\vspace{-2mm}
	\setlength\tabcolsep{8pt}
	\scalebox{1}[1]{
	\begin{tabular}{cccccc}
		\toprule
		\multicolumn{1}{c}{\multirow{3}{*}{Dataset}}
		& \multicolumn{1}{c}{\multirow{3}{*}{Metrics}}
		& \multicolumn{4}{c}{$\lambda_2$}  \\
		\cmidrule(l){3-6}
		
		&  & $\lambda_2$ = 1   & $\lambda_2$ = 5   & $\lambda_2$ = 10  & $\lambda_2$ = 15  \\
		
		\cmidrule(l){1-6}
		\multicolumn{1}{c}{\multirow{2}{*}{SOTS-Indoor}} & PSNR  &41.48  &40.28   &\textbf{41.92}   &39.91     \\
		\multicolumn{1}{c}{}                             & SSIM  &0.9952 &0.9935  &\textbf{0.9954}  &0.9932     \\
		\bottomrule
	\end{tabular}}
	\label{BetaAnalysis}
\end{table}

\begin{figure}[t!]
	\vspace{-4mm}
	\scriptsize
	\centering
	\subfigure{}
	\vspace{-4mm}
	\subfigure{
		\rotatebox[]{90}{\scriptsize{Input}}
		\hspace{1pt}
		\begin{minipage}{0.31\linewidth}
			\vspace{3pt}
			\centerline{\includegraphics[width=2.6cm,height=1.8cm]{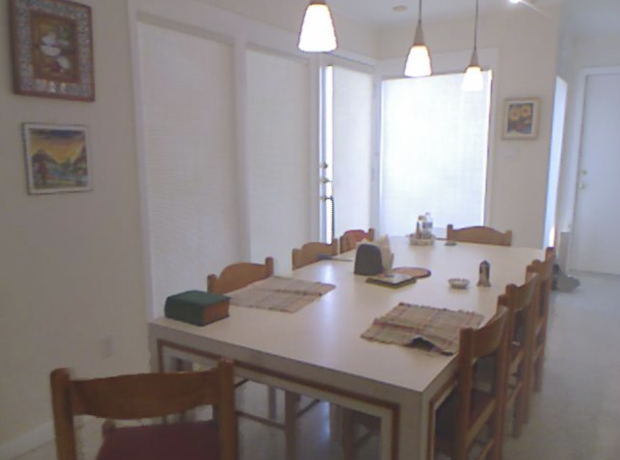}}
		\end{minipage}
		\begin{minipage}{0.31\linewidth}
			\vspace{3pt}
			\centerline{\includegraphics[width=2.6cm,height=1.8cm]{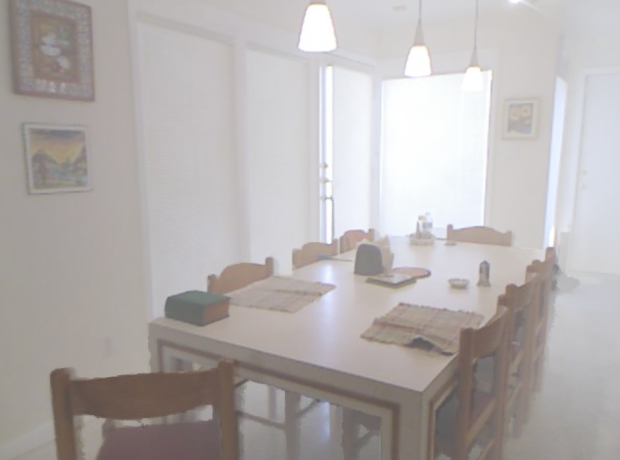}}
		\end{minipage}
		\begin{minipage}{0.31\linewidth}
			\vspace{3pt}
			\centerline{\includegraphics[width=2.6cm,height=1.8cm]{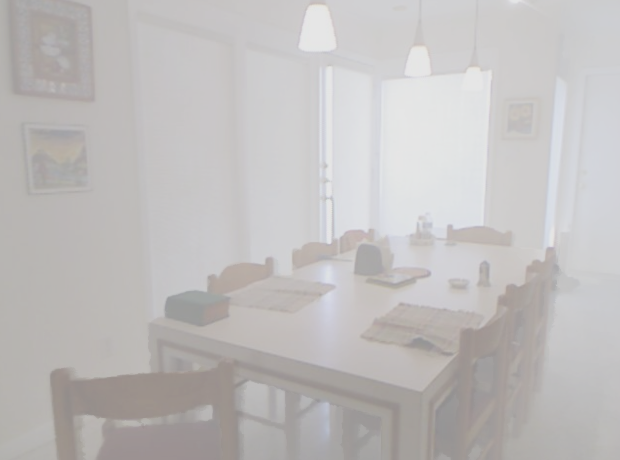}}
		\end{minipage}}
	\vspace{-4mm}
	\subfigure{
		\rotatebox[]{90}{\scriptsize{DCP}}
		\hspace{1pt}
		\begin{minipage}{0.31\linewidth}
			\vspace{3pt}
			\centerline{\includegraphics[width=2.6cm,height=1.8cm]{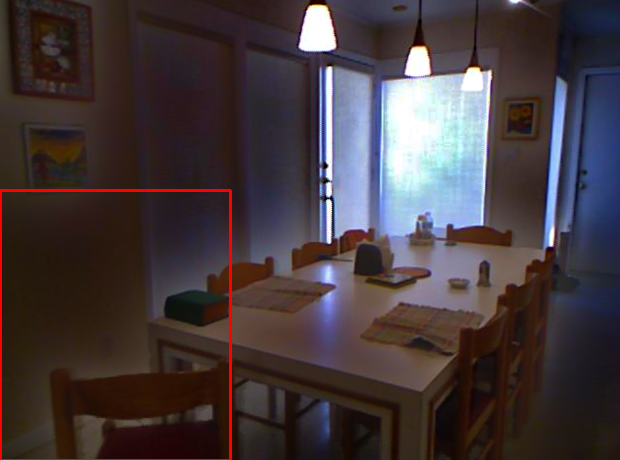}}
		\end{minipage}
		\begin{minipage}{0.31\linewidth}
			\vspace{3pt}
			\centerline{\includegraphics[width=2.6cm,height=1.8cm]{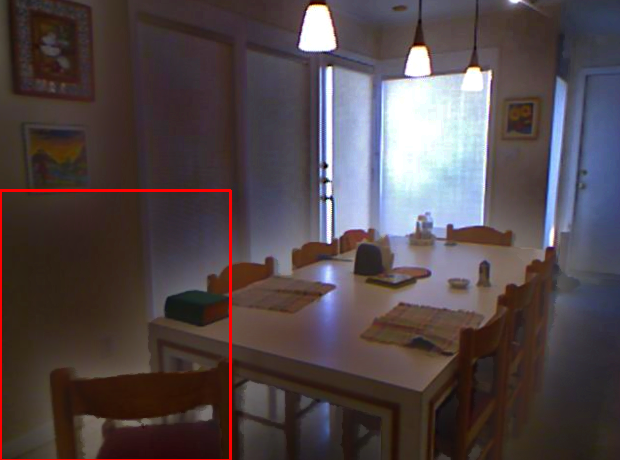}}
		\end{minipage}
		\begin{minipage}{0.31\linewidth}
			\vspace{3pt}
			\centerline{\includegraphics[width=2.6cm,height=1.8cm]{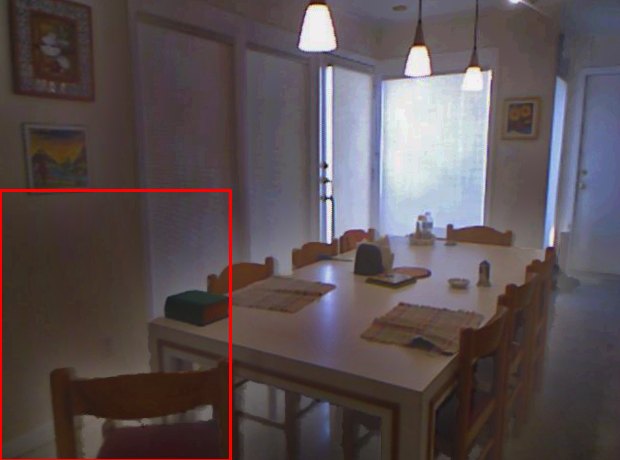}}
	\end{minipage}}
	\vspace{-4mm}
	\subfigure{
		\rotatebox[]{90}{\scriptsize{Grid}}
		\hspace{1pt}
		\begin{minipage}{0.31\linewidth}
			\vspace{3pt}
			\centerline{\includegraphics[width=2.6cm,height=1.8cm]{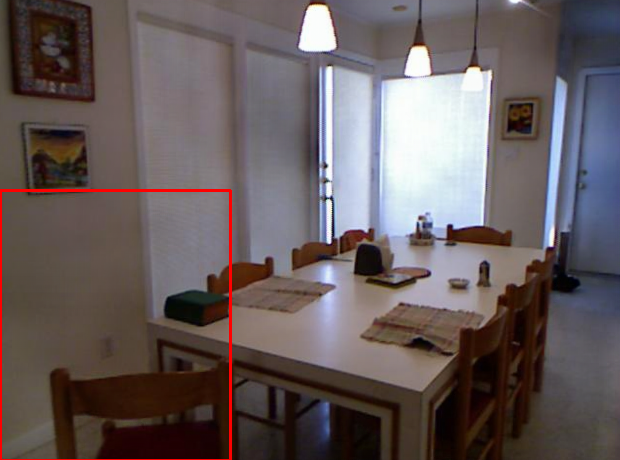}}
		\end{minipage}
		\begin{minipage}{0.31\linewidth}
			\vspace{3pt}
			\centerline{\includegraphics[width=2.6cm,height=1.8cm]{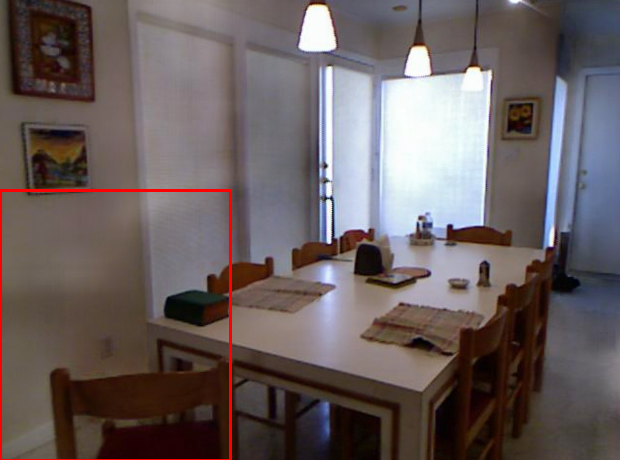}}
		\end{minipage}
		\begin{minipage}{0.31\linewidth}
			\vspace{3pt}
			\centerline{\includegraphics[width=2.6cm,height=1.8cm]{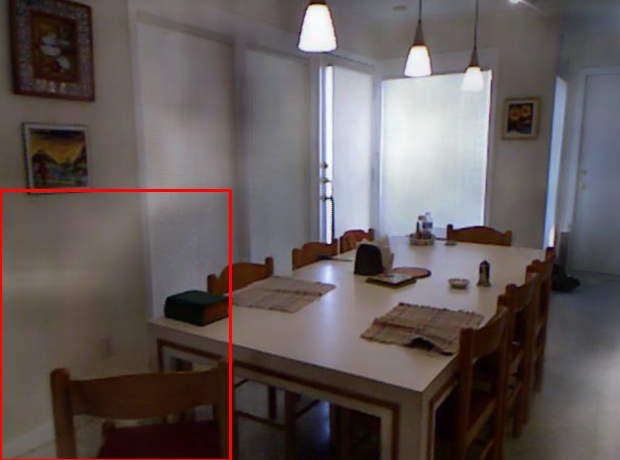}}
	\end{minipage}}
	\vspace{-4mm}
	\subfigure{
		\rotatebox[]{90}{\scriptsize{FFA}}
		\hspace{1pt}
		\begin{minipage}{0.31\linewidth}
			\vspace{3pt}
			\centerline{\includegraphics[width=2.6cm,height=1.8cm]{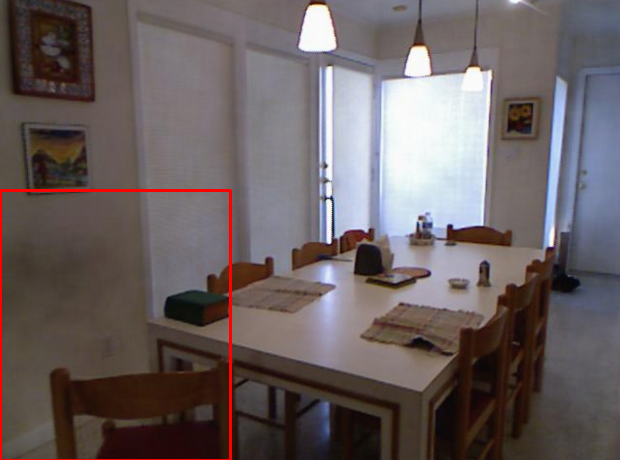}}
		\end{minipage}
		\begin{minipage}{0.31\linewidth}
			\vspace{3pt}
			\centerline{\includegraphics[width=2.6cm,height=1.8cm]{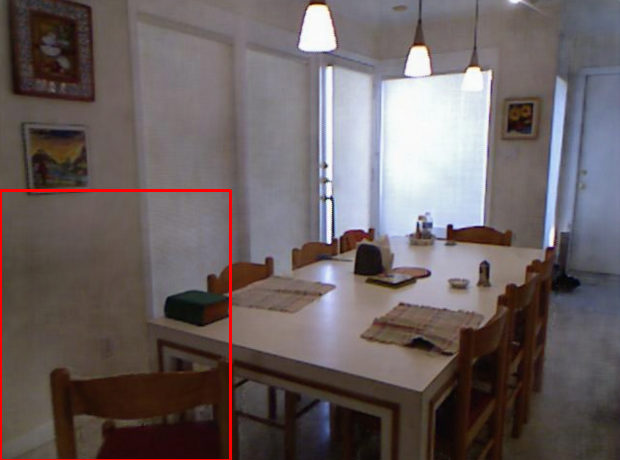}}
		\end{minipage}
		\begin{minipage}{0.31\linewidth}
			\vspace{3pt}
			\centerline{\includegraphics[width=2.6cm,height=1.8cm]{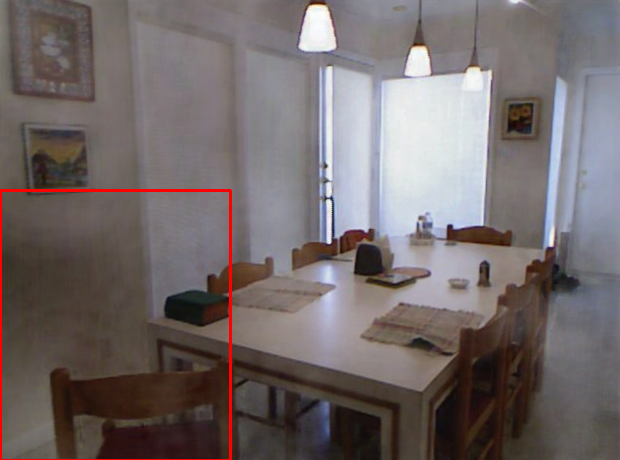}}
	\end{minipage}}
	\vspace{-4mm}
	\subfigure{
		\rotatebox[]{90}{\scriptsize{Ours}}
		\hspace{1pt}
		\begin{minipage}{0.31\linewidth}
			\vspace{3pt}
			\centerline{\includegraphics[width=2.6cm,height=1.8cm]{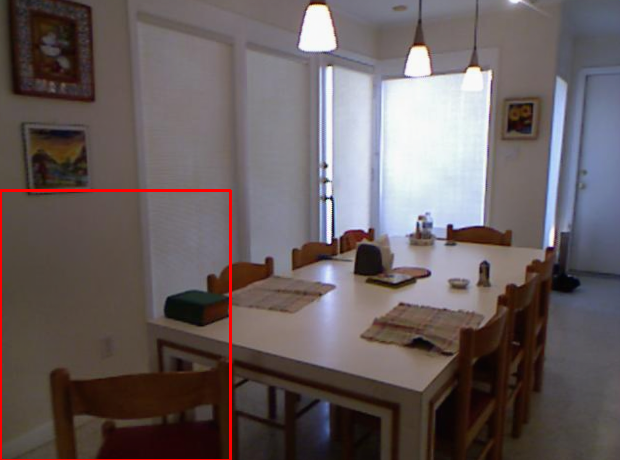}}
			\centerline{Mild-level}
		\end{minipage}
		\begin{minipage}{0.31\linewidth}
			\vspace{3pt}
			\centerline{\includegraphics[width=2.6cm,height=1.8cm]{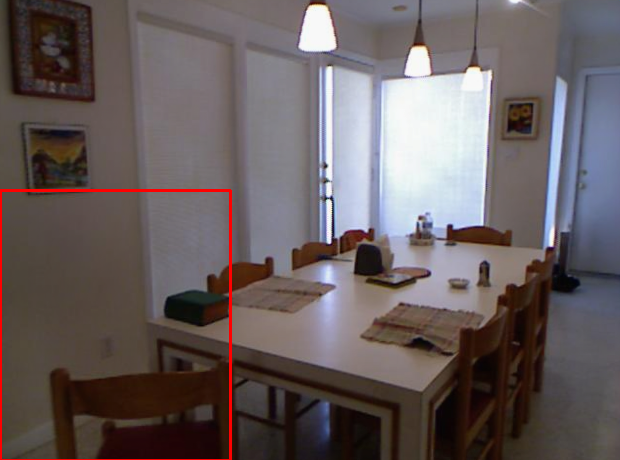}}
			\centerline{Medium-level}
		\end{minipage}
		\begin{minipage}{0.31\linewidth}
			\vspace{3pt}
			\centerline{\includegraphics[width=2.6cm,height=1.8cm]{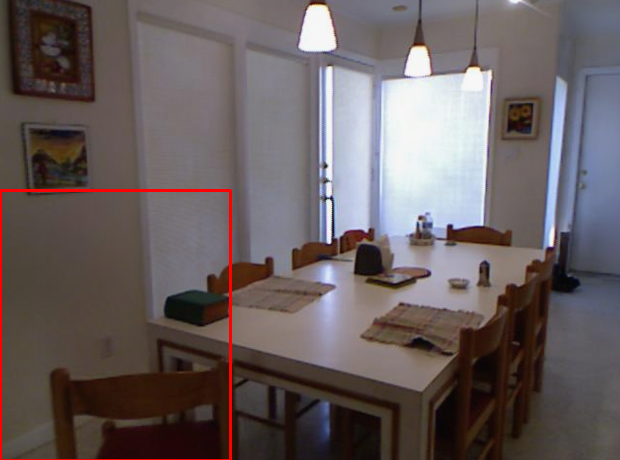}}
			\centerline{Heavy-level}
	\end{minipage}}
	\caption{The dehazed results of different methods on the image with different hazy densities.}
	\label{different_haze}
	\vspace{-3mm}
\end{figure}

\subsection{Ablation Study}
The proposed image dehazing method contains two novel ingredients: 1) The proposed CARL to better optimize the network parameters; 2) The consistency regularization framework to further improve the model robustness. To reveal how each ingredient contributes to the performance improvement, we conduct ablation study to analyze different elements, including the $\mathcal{L}_1$, $\mathcal{L}_{CR}$ and $\mathcal{L}_{CARL}$, on both synthetic and real-world hazy image datasets.

We conduct all the experiments based on the same dehazing network architecture (FFA-Net~\cite{qin2020ffa}). We implement the following four variants of the proposed method: 1) $\mathcal{L}_1$: Training the network by the traditional $L_1$ loss function, which works as the baseline method; 2) $\mathcal{L}_1$+$\mathcal{L}_{CARL}$: Training the network jointly with the $L_1$ loss and our proposed CARL; 3) $\mathcal{L}_1$+$\mathcal{L}_{DivC}$~\cite{wu2021contrastive}: Training the network jointly with the $L_1$ loss and the divided-contrast loss~\cite{wu2021contrastive}, which is to make a comparison with ours; 4) $\mathcal{L}_1$+$\mathcal{L}_{CARL}$ + $\mathcal{L}_{CR}$: Training the network jointly with the $L_1$ loss, CARL and the consistency regularization, which is our final algorithm as illustrated in Eq.~\ref{Eq:OverallLoss}.

The performance of these models are summarized in Table~\ref{Analysis}, we can clearly see that by adding our proposed CARL into the traditional $L_1$ loss, we can improve the baseline performance 3.17db and 0.91db PSNR on the SOTS-Indoor and Dense-Haze datasets, respectively. Compared with the relevant method $\mathcal{L}_{DivC}$~\cite{wu2021contrastive}, our proposed method outperforms $\mathcal{L}_{DivC}$ by a margin of 2.02db and 0.48db PSNR on these two  datasets, respectively. When further adding the proposed consistency regularization $\mathcal{L}_{CR}$, we can also improve the performance by 2.36db and 0.17db PSNR on these two datasets respectively. As shown in Figure~\ref{different_haze}, we can clearly see that our method generates more consistent dehazed images, and can deal with multi-level hazy images well.
Therefore, such detailed experimental results greatly illustrate the effectiveness of our proposed method.

\textbf{Parameter sensitivity analysis} is shown in Table~\ref{BetaAnalysis}. As defined in Eq.~\ref{Eq:OverallLoss}, our final loss function contains three terms: $\mathcal{L}_{1}$, $\mathcal{L}_{CR}$  and $\mathcal{L}_{CARL}$. To investigate the effect of hyper-parameters on the performance, we conduct comprehensive experiments with various values of these parameters. Here, we just list the performance with various $\lambda_2$ when $\lambda_1=1.0$ in Table~\ref{BetaAnalysis}, for the limitation of paper length. We can clearly see that our method yields best performance when $\lambda_2=10.0$.

\vspace{-2mm}
\section{Conclusion}
In this paper, we propose a contrast-assisted reconstruction loss for single image dehazing. The proposed method can fully
exploit the negative information to better facilitate the traditional positive-orient dehazing objective function. Besides, we also propose the consistency regularization to further improve the model robustness and consistency. The proposed method can work as a universal learning framework to further improve the performance of various state-of-the-art dehazing networks without introducing additional computation/parameters during testing phase. In the future, we will extend our method to many other relevant tasks, such as image deraining, image super-resolution, etc.

\bibliographystyle{named}
\bibliography{ijcai22}

\end{document}